\documentclass{isprs} 
\usepackage{subcaption}
\usepackage{setspace}
\usepackage{geometry} 
\usepackage{epstopdf}
\usepackage[labelsep=period]{caption}  
\usepackage[british]{babel} 
\usepackage[hang]{footmisc}
\usepackage{booktabs}
\usepackage{amsmath}
\usepackage{multirow}
\usepackage{natbib}
\usepackage{amssymb}

\begin{document}

\title{The Impact of CutMix on Reliability and Robustness in Semantic Segmentation}
\date{}


\author{Steven Landgraf \thanks{Corresponding Authors} , Markus Ulrich}

\address{Institute of Photogrammetry and Remote Sensing (IPF),\\ Karlsruhe Institute of Technology (KIT), Germany,\\ (steven.landgraf, markus.ulrich)@kit.edu}



\abstract{
Ensuring not only high accuracy but also reliable and robust predictions is critical for the deployment of semantic segmentation models in safety-critical applications such as autonomous driving. Despite the widespread use of CutMix -- a simple yet powerful data augmentation strategy -- its effect on the reliability and robustness in dense predictions tasks remains unexplored. Motivated by recent findings that semi-supervised segmentation methods, where CutMix is a core component, can severely degrade reliability, this study isolates and systematically analyzes the influence of CutMix on segmentation accuracy, calibration, and uncertainty quality. We evaluate two representative architectures, the CNN-based DeepLabV3+ and the transformer-based SegFormer, across both in-domain and out-of-domain scenarios. Our results show that CutMix has only a minor impact on segmentation accuracy but consistently improves the reliability, particularly under distribution shifts. These improvements indicate that CutMix primarily enhances the trustworthiness of the model's calibration and uncertainty rather than the raw segmentation prediction itself. This distinction is crucial for safety-critical deployment, where reliable confidence estimates are as important as raw performance.
}


\maketitle

\section{Introduction}\label{MANUSCRIPT}
\sloppy
Deep neural networks achieve remarkable success in semantic segmentation \citep{minaee2021image}, driving progress in a variety of domains ranging from autonomous driving, medical imaging, and remote sensing \citep{muhammad2022vision,azad2024medical,li2024review}. However, as their deployment in safety-critical real-world applications increases, so does the need to ensure that their predictions are not only accurate but also reliable and robust. Reliability, in this context, refers to the degree to which model confidences reflect the true likelihood of correctness, while robustness describes the model’s ability to maintain performance under perturbations or distribution shifts.

Due to the cost -- and therefore scarcity -- of manually labeled ground truth information for semantic segmentation tasks, recent work has embraced semi-supervised learning and strong data augmentations as core design elements. Among these, CutMix \citep{yun2019cutmix} has become a central component in recent semi-supervised learning frameworks \citep{yang2023revisiting,yang2025unimatch}. By randomly pasting patches from one image into another and mixing their corresponding labels, CutMix promotes spatially localized feature learning and reduces overfitting. While this simple, yet effective strategy has been shown to improve classification reliability and robustness \citep{oh2024provable,rao2023studying,yun2019cutmix}, its impact on dense prediction tasks such as semantic segmentation remain poorly understood.

In addition, a recent study revealed a critical blind spot in this context: state-of-the-art semi-supervised segmentation methods, despite their strong performance in terms of accuracy, can severely deteriorate the reliability of neural networks \citep{landgraf2025rethinking}. This observation raises an important research question, as CutMix is a key component of these methods and may contribute to this problem. 

Motivated by these findings, this work aims to analyze the impact of CutMix on reliability and robustness in semantic segmentation. We evaluate the effects of CutMix training in both in-domain and out-of-domain scenarios, considering accuracy, calibration, uncertainty quality, and their robustness. By disentangling the influence of CutMix from other components of semi-supervised learning frameworks, we aim to reveal its impact on the reliability and robustness in semantic segmentation -- a research question that has yet to be answered.  

\section{Related Work}\label{RELATED_WORK}

\begin{figure*}[!ht]
    \centering
    \begin{subfigure}{0.2\textwidth}
        \includegraphics[width=\textwidth]{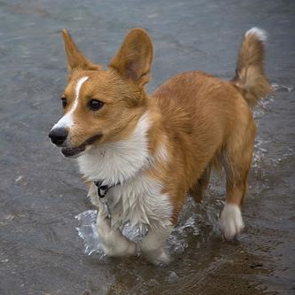}
        \caption{Input Image \\Label: Dog 1.0}
    \end{subfigure}
    \begin{subfigure}{0.2\textwidth}
        \includegraphics[width=\textwidth]{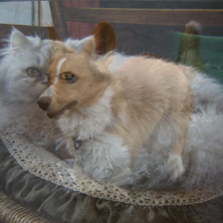}
        \caption{MixUp \\Label: Dog 0.5, Cat 0.5}
    \end{subfigure}
    \begin{subfigure}{0.2\textwidth}
        \includegraphics[width=\textwidth]{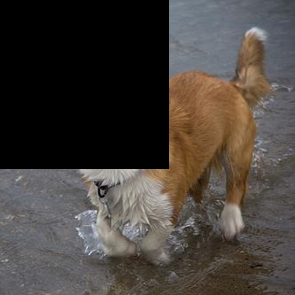}
        \caption{CutOut \\Label: Dog 1.0}
    \end{subfigure}
    \begin{subfigure}{0.2\textwidth}
        \includegraphics[width=\textwidth]{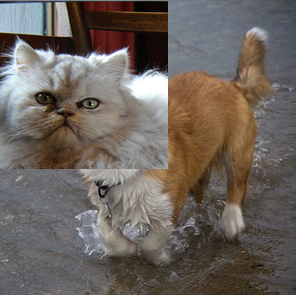}
        \caption{CutMix \\Label: Dog 0.6, Cat 0.4}
    \end{subfigure}
    \caption{A comparison between MixUp \citep{zhang2017mixup}, CutOut \citep{devries2017improved}, and CutMix \citep{yun2019cutmix}. Images taken from \citet{yun2019cutmix}.}
    \label{fig: comparison}
\end{figure*}

\textbf{Reliability.} A model's reliability encompasses its calibration and uncertainty quality. The former describes how well the confidences of the predicted class reflect the true likelihood of correctness, most commonly measured by the Expected Calibration Error (ECE) \citep{guo2017calibration}. A model's uncertainty quality describes its ability to align the entire softmax output with predictive ambiguities and errors \citep{mukhoti2018evaluating}.

Despite impressive predictive capabilities, neural networks are known to suffer from poor calibration \citep{guo2017calibration,wilson2020bayesian,wang2021rethinking}. In light of this, \citet{guo2017calibration} have introduced temperature scaling as a straightforward, yet effective post-hoc calibration method. Because of its simplicity and non-invasiveness it is still widely used as a baseline for numerous alternative approaches \citep{kull2019beyond,naeini2015obtaining,ji2019bin,ding2021local,patra2023calibrating}. Besides calibration, a variety of techniques have been proposed to estimate high-quality uncertainties in deep learning models \citep{mackay1992practical,gal2016dropout,lakshminarayanan2017simple,valdenegro2023sub,van2020uncertainty,liu2020simple,mukhoti2023deep,amini2020deep,landgraf2024dudes,landgraf2025efficient}. Unfortunately, however, all of these either introduce technical complexity or induce high computational cost, making them impractical for real-world applications like autonomous driving \citep{muhammad2020deep}. 

\textbf{Robustness.} Whilst all of the previously mentioned methods offer an effective way of enhancing reliability in in-domain settings, there is no guarantee of the generalizability to out-of-domain scenes. In fact, multiple prior works have found that reliability deteriorates significantly under domain shifts \citep{ovadia2019can,de2023reliability,landgraf2025comparative}. Consequently, model robustness, i.e., its ability to remain effective under perturbations, noise, or distribution shifts, is attracting increasing attention. Prior work discerns between robustness to perturbations (Gaussian noise, blur, occlusion), adversarial attacks (imperceptible changes to an image crafted to induce failure), and natural domain shifts (variations in weather, lighting, or geographic context) \citep{hendrycks2019benchmarking,kamann2020benchmarking,kamann2021benchmarking,goodfellow2014explaining,hendrycks2021natural,pedraza2022really,recht2019imagenet,sakaridis2021acdc,hu2019depth,sakaridis2018semantic,varma2019idd}. 

\textbf{Research Gap.} By virtue of the importance of this topic, we are not the first to analyze reliability and robustness of deep learning models in the context of semantic segmentation \citep{arnab2018robustness,kamann2020benchmarking,kamann2021benchmarking,zhou2022understanding,de2023reliability,loiseau2024reliability,zhou2019automated}. However, a recent study by \citet{landgraf2025rethinking} revealed a critical blind spot, showing that state-of-the-art semi-supervised segmentation methods severely deteriorate the reliability of neural networks. A core component in these methods is CutMix \citep{yun2019cutmix}, which has been shown to have provable benefit for feature learning and improved reliability and robustness in classification tasks \citep{oh2024provable,rao2023studying,yun2019cutmix}. At the same time, there is no clear answer as to how CutMix impacts reliability and robustness in dense prediction models, leaving a gap that this work aims to address.

\section{Experimantal Setup}\label{METHODOLOGY}
The following describes the methodological background of this study, our training configurations, evaluation metrics, and datasets used to analyze the impact of CutMix on reliability and robustness in semantic segmentation. 

\subsection{CutMix}
\begin{figure}[!t]
    \centering
    \begin{subfigure}{0.23\textwidth}
        \includegraphics[width=\textwidth]{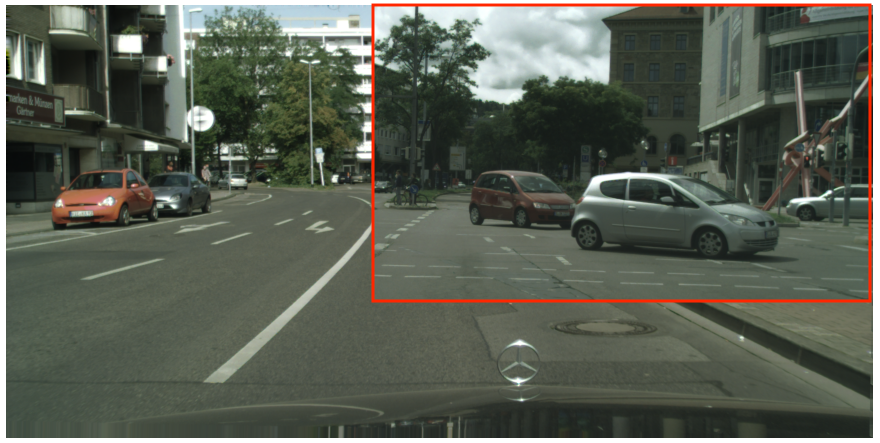}
        \caption{Input Image}
    \end{subfigure}
    \begin{subfigure}{0.23\textwidth}
        \includegraphics[width=\textwidth]{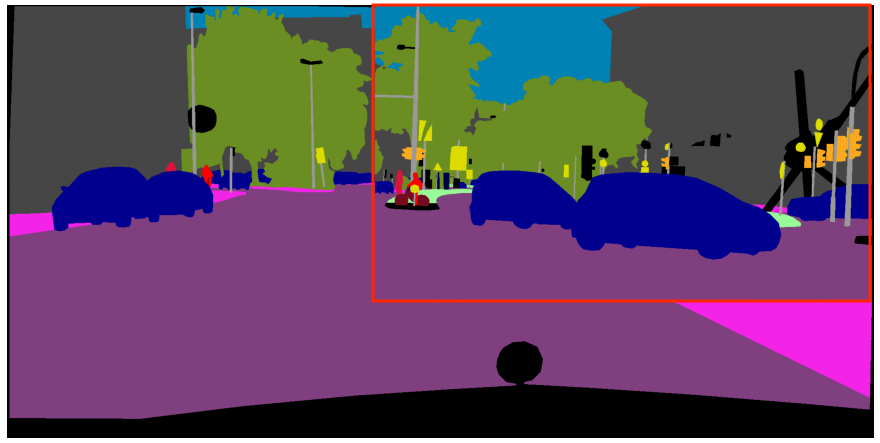}
        \caption{Ground Truth}
    \end{subfigure}
    \caption{CutMix training example on the Cityscapes dataset \citep{cordts2016cityscapes}. For visualization purposes, we have highlighted the inserted area with a red rectangle.}
    \label{fig: cutmix}
\end{figure}

CutMix \citep{yun2019cutmix} is highly inspired by both MixUp \citep{zhang2017mixup} and CutOut \citep{devries2017improved}, as shown by Figure \ref{fig: comparison}. The former blends a pair of images and labels based on convex combinations, while the latter randomly masks out regions of the input to regularize the model. CutMix combines both of these ideas by randomly cutting and pasting patches among training images while proportionally mixing the ground truth labels, as shown by Figure \ref{fig: cutmix}.

Formally, a CutMix training sample $(\tilde{x},\tilde{y})$ can be defined as 
\begin{align}
&\tilde{x} = M \odot x_A + (1 - M) \odot x_B \enspace, \\
&\tilde{y} = \lambda y_A + (1 - \lambda) y_B \enspace,
\end{align}
where $x_A$ and $x_B$ denote two input images, and $y_A$ and $y_B$ their corresponding one-hot encoded labels. The binary mask $M \in \{0,1\}^{HxW}$ specifies a randomly sampled rectangular region within the image, determining which pixels are taken from image $A$ and which from image $B$. The mixing ratio $\lambda$ represents the proportion of pixels retained from $x_A$ and is sampled from the beta distribution. With respect to the hyperparameters, we follow the original implementation \citep{yun2019cutmix}. 

\subsection{Training Configuration}
All experiments were conducted using two widely adopted semantic segmentation architectures: DeepLabV3+ (DLV3+) \citep{Chen_2018_ECCV} and SegFormer \citep{xie2021segformer}. Both models were trained for 250 epochs without early stopping and with a batch size of 8 using a polynomial learning rate schedule. The initial learning rates and weight decay were selected following the recommended configurations for each architecture, as summarized by Table~\ref{tab:hyperparams}. All models were optimized using the standard pixel-wise cross-entropy loss and AdamW optimizer \citep{loshchilov2017decoupled} with standard momentum parameters. These consistent hyperparameter and augmentation settings were chosen to ensure comparability between architectures and to isolate the specific effects of CutMix on reliability and robustness -- both for Convolutional Neural Networks and Vision Transformers.

\begin{table*}
    \centering
    \begin{tabular}{l|c|c|c|c|c}
        Model & Learning Rate (LR) & Weight Decay & Epochs & Batch Size & LR Schedule \\
        \midrule \midrule
        DeepLabV3+ & $1\times10^{-4}$ ($\times$10 Decoder) & $1\times10^{-4}$ & \multirow{2}{*}{250} & \multirow{2}{*}{8} & \multirow{2}{*}{Polynomial} \\
        SegFormer & $5\times10^{-6}$ ($\times$10 Decoder) & $1\times10^{-2}$ & & & \\
    \end{tabular}
    \caption{Training hyperparameters for DeepLabV3+ and SegFormer.}
    \label{tab:hyperparams}
\end{table*}

\subsection{Metrics}
We evaluate model performance using a combination of accuracy, calibration, and uncertainty-based metrics. The mean Intersection over Union (mIoU) \citep{lateef2019survey} serves as the primary measure of segmentation accuracy. Model calibration is assessed by the Expected Calibration Error (ECE) \citep{guo2017calibration}, which quantifies the discrepancy between predictive confidence and empirical accuracy. To capture the quality of uncertainty estimates, we first compute the pixel-wise predictive entropy \citep{shannon1948mathematical}:
\begin{equation}\label{eq: entropy}
H(x) = -\sum_{c=1}^C p(\hat{y}_c(x)) \log p(\hat{y}_c(x))\enspace,
\end{equation}
where $p(\hat{y}_c(x))$ denotes the predicted softmax probability for class $c$ given an input image $x$. These entropy values are then used to classify pixels as “certain” or “uncertain” based on the median uncertainty within each image, which was found to be the best default threshold \citep{landgraf2025comparative}. The conditional metrics p(\text{acc}$|$\text{cer}) and p(\text{unc}$|$\text{inacc}) proposed by \citet{mukhoti2018evaluating} then quantify how often predictions are correct when marked certain, and incorrect when marked uncertain, respectively. 

The Reliable Segmentation Score (RSS) \citep{landgraf2025comparative} integrates all previous complementary metrics into a single, holistic reliability measure using the harmonic mean:
\begin{equation}\label{eq: rss}
    \text{RSS} = \frac{\sum \omega_i}{\frac{\omega_1}{\text{mIoU}} + \frac{\omega_2}{(1-\text{ECE})} + \frac{\omega_3}{\text{p(acc$|$cer)}} + \frac{\omega_3}{\text{p(unc$|$inacc)}}}\enspace,
\end{equation}
where the application-specific weights $\omega_i$ are all set to $1.0$ to avoid assumptions about the importance of any one metric. By leveraging the harmonic mean, RSS penalizes poor performance in any aspect, ensuring that a model achieves a high score only if it is accurate, well-calibrated, and uncertainty-aware. 

The combination of these metrics is justified by their largely orthogonal contributions \citep{landgraf2025comparative}. While the mIoU captures pixel-wise accuracy across classes, the ECE
measures calibration, i.e., the alignment between predicted confidence and true likelihood of correctness, which can vary independently of mIoU. p(acc$|$cer) evaluates whether low-uncertainty predictions are indeed correct, highlighting the usefulness of certainty. p(unc$|$inacc) assesses whether false predictions are flagged as uncertain, enabling error mitigation. Besides, unlike mIoU and ECE, which rely solely on the maximum softmax probability and therefore capture information about the predicted class only, the conditional uncertainty metrics consider the full softmax distribution (via entropy, see Eq. \ref{eq: entropy}), providing insights into prediction ambiguity.

\subsection{Datasets}
To evaluate the impact of CutMix training for real-world scenarios, we train all models on Cityscapes \citep{cordts2016cityscapes}. For out-of-domain analyses, we use the Foggy Cityscapes \citep{sakaridis2018semantic} validation sets without re-training the models. Foggy Cityscapes provides three versions defined by the attenuation coefficient $\beta$, where higher values correspond to denser fog. This setup enables a systematic assessment of model reliability and robustness under progressively challenging visual degradations.

In terms of data augmentations, we employed random scaling, horizontal flipping, and random cropping as a baseline for all models. When evaluating the effect of CutMix, we applied it with a probability of 50\%, following the original formulation \citep{yun2019cutmix}.

\section{Results}\label{EXPERIMENTS}
\textbf{In-Domain Evaluation.} 
Table~\ref{tab:id_evaluation} summarizes the in-domain results on Cityscapes for different model backbones, comparing standard training and CutMix-augmented variants. As expected, larger backbones achieve higher segmentation accuracy (mIoU), while calibration (ECE) remains relatively stable across architectures. Interestingly, uncertainty quality -- as measured by p(\text{acc}$|$\text{cer}) and p(\text{unc}$|$\text{inacc}) -- appears largely independent of backbone size, with the CNN-based DeepLabV3+ slightly outperforming the Vision Transformer-based SegFormer models. Runtime scales predictably with model complexity, from $\approx$ 25 ms (40 FPS) for DeepLabV3+ (ResNet-34) to $\approx$ 315 ms (3 FPS) for SegFormer (B5). While the transformer-based SegFormer models offer competitive accuracy, they come at a considerably higher computational cost, emphasizing the continued efficiency advantage of CNN-based designs, particularly for images with high resolutions.

Overall, CutMix has a minor effect on segmentation accuracy and calibration, improving mIoU in three out of six cases and leaving ECE nearly unchanged. However, it consistently enhances uncertainty quality, particularly in p(\text{unc}$|$\text{inacc}), where most models show notable gains. The only exception is the largest SegFormer (MiT-B5), whose reported checkpoint underperformed despite showing better averages across training epochs, clearly a case of suboptimal checkpoint selection. However, to ensure consistency, we retained this checkpoint rather than retraining or using early stopping. When aggregating evaluation results with the Reliable Segmentation Score (RSS), CutMix-trained models outperform their counterparts in all of the remaining models, indicating that CutMix substantially improves reliability. Surprisingly, the older CNN-based DeepLabV3+ architecture remains more reliable overall, surpassing the modern transformer-based models in terms of uncertainty quality.

\begin{table*}[!h]
    \centering
    \setlength{\tabcolsep}{7pt} 
    \renewcommand{\arraystretch}{1.1}
    \begin{tabular}{l|c|c|c|c|c|c|c|c|c}
        Encoder & Params & CutMix & mIoU $\uparrow$ 
        & ECE $\downarrow$ 
        & $p(\text{acc}|\text{cer})$ $\uparrow$ 
        & $p(\text{unc}|\text{inacc})$ $\uparrow$ 
        & RSS $\uparrow$ & Inference Time [ms] $\downarrow$ & FPS $\uparrow$ \\
        \midrule \midrule
        \multicolumn{10}{c}{\textbf{DeepLabV3+}} \\
        \midrule \midrule
        \multirow{2}{*}{RN34}   & \multirow{2}{*}{$\sim$ 21M} &      & 0.743 & 0.033 & 0.912 & 0.731 & 0.826 & \multirow{2}{*}{24.81 $\pm$ 8.50} & \multirow{2}{*}{40.31} \\
           &                        & \checkmark                     & 0.743 & \textbf{0.027} & \textbf{0.943} & \textbf{0.838} & \textbf{0.864} & \\ \midrule
        \multirow{2}{*}{RN101}  & \multirow{2}{*}{$\sim$ 42M} &      & \textbf{0.774} & 0.034 & 0.931 & 0.797 & 0.859 & \multirow{2}{*}{49.72 $\pm$ 11.01} & \multirow{2}{*}{20.11}\\
               &                        & \checkmark                 & 0.754 & \textbf{0.033} & \textbf{0.947} & \textbf{0.848} & \textbf{0.870} & \\ \midrule
        \multirow{2}{*}{RN152}  & \multirow{2}{*}{$\sim$ 58M} &      & 0.762 & 0.034 & 0.927 & 0.777 & 0.849 & \multirow{2}{*}{57.31 $\pm$ 15.43} & \multirow{2}{*}{22.88} \\
               &                        & \checkmark                 & \textbf{0.774} & 0.034 & \textbf{0.939} & \textbf{0.817} & \textbf{0.867} & \\
        \midrule \midrule
        \multicolumn{10}{c}{\textbf{SegFormer}} \\
        \midrule \midrule
        \multirow{2}{*}{MiT-B0} & \multirow{2}{*}{$\sim$ 3.7M}  &      & 0.658 & 0.034 & 0.899 & 0.713 & 0.789 & \multirow{2}{*}{43.71 $\pm$ 0.24} & \multirow{2}{*}{22.88} \\
               &                        & \checkmark                   & \textbf{0.688} & 0.034 & \textbf{0.930} & \textbf{0.807} & \textbf{0.833} & \\ \midrule
        \multirow{2}{*}{MiT-B3} & \multirow{2}{*}{$\sim$ 45M}   &      & 0.759 & \textbf{0.032} & 0.906 & 0.707 & 0.822 & \multirow{2}{*}{181.11 $\pm$ 0.71} & \multirow{2}{*}{5.52} \\
               &                        & \checkmark                   & \textbf{0.771} & 0.034 & \textbf{0.921} & \textbf{0.758} & \textbf{0.844} & \\ \midrule
        \multirow{2}{*}{MiT-B5} & \multirow{2}{*}{$\sim$ 82M}   &      & \textbf{0.788} & \textbf{0.027} & \textbf{0.921} & \textbf{0.766} & \textbf{0.853} & \multirow{2}{*}{315.50 $\pm$ 37.27} & \multirow{2}{*}{3.17} \\
               &                        & \checkmark                   & 0.773 & 0.033 & 0.909 & 0.721 & 0.831 & \\
       \bottomrule
    \end{tabular}
    \caption{In-domain evaluation results on the Cityscapes validation dataset using DeepLabV3+ and SegFormer models with different backbone configurations. The reported inference times and frames per second (FPS) correspond to single-image forward passes performed at the native Cityscapes resolution (1024$\times$2048) without any inference-time optimizations such as mixed precision or batching. All measurements were conducted on a single NVIDIA A100 GPU to ensure a consistent and comparable runtime evaluation across model architectures.}
    \label{tab:id_evaluation}
\end{table*}

\begin{table*}[!h]
    \centering
    \setlength{\tabcolsep}{3pt}
    \renewcommand{\arraystretch}{1.25}
    \begin{tabular}{
        l|c|ccc|ccc|ccc|ccc|ccc
    }
        \multirow{2}{*}{Encoder} & \multirow{2}{*}{CutMix} &
        \multicolumn{3}{c|}{mIoU $\uparrow$} &
        \multicolumn{3}{c|}{ECE $\downarrow$} &
        \multicolumn{3}{c|}{$p(\text{acc}|\text{cer})$ $\uparrow$} &
        \multicolumn{3}{c|}{$p(\text{unc}|\text{inacc})$ $\uparrow$} &
        \multicolumn{3}{c}{RSS $\uparrow$} \\ 
        \cline{3-5} \cline{6-8} \cline{9-11} \cline{12-14} \cline{15-17}
         &  & Fog$_1$ & Fog$_2$ & Fog$_3$ & Fog$_1$ & Fog$_2$ & Fog$_3$ & Fog$_1$ & Fog$_2$ & Fog$_3$ & Fog$_1$ & Fog$_2$ & Fog$_3$ & Fog$_1$ & Fog$_2$ & Fog$_3$ \\
        \midrule \midrule
        \multicolumn{17}{c}{\textbf{DeepLabV3+}} \\
        \midrule \midrule
        \multirow{2}{*}{RN34} & & 0.702 & 0.651 & 0.568 & 0.041 & 0.067 & 0.092 & 0.907 & 0.901 & 0.886 & 0.741 & 0.753 & 0.753 & 0.813 & 0.793 & 0.752 \\
                              & \checkmark & 0.701 & 0.654 & 0.567 & 0.038 & 0.068 & 0.069 & 0.941 & 0.934 & 0.913 & 0.851 & 0.851 & 0.845 & 0.850 & 0.825 & 0.782 \\ \midrule
        \multirow{2}{*}{RN101} & & 0.734 & 0.682 & 0.585 & 0.046 & 0.072 & 0.107 & 0.927 & 0.920 & 0.903 & 0.803 & 0.810 & 0.821 & 0.845 & 0.822 & 0.776 \\
                               & \checkmark & 0.726 & 0.675 & 0.565 & 0.038 & 0.076 & 0.081 & 0.941 & 0.932 & 0.909 & 0.845 & 0.842 & 0.836 & 0.858 & 0.829 & 0.776 \\ \midrule
        \multirow{2}{*}{RN152} & & 0.726 & 0.679 & 0.573 & 0.044 & 0.068 & 0.120 & 0.921 & 0.908 & 0.890 & 0.782 & 0.773 & 0.798 & 0.835 & 0.810 & 0.761 \\
                               & \checkmark & 0.740 & 0.696 & 0.605 & 0.033 & 0.058 & 0.074 & 0.936 & 0.928 & 0.911 & 0.823 & 0.816 & 0.818 & 0.857 & 0.833 & 0.792 \\
        \midrule \midrule
        \multicolumn{17}{c}{\textbf{SegFormer}} \\
        \midrule \midrule
        \multirow{2}{*}{MiT-B0} & & 0.616 & 0.566 & 0.470 & 0.060 & 0.043 & 0.036 & 0.918 & 0.914 & 0.890 & 0.801 & 0.819 & 0.819 & 0.796 & 0.780 & 0.726 \\
                                & \checkmark & 0.640 & 0.578 & 0.467 & 0.048 & 0.031 & 0.022 & 0.936 & 0.927 & 0.890 & 0.852 & 0.857 & 0.836 & 0.824 & 0.799 & 0.729 \\ \midrule
        \multirow{2}{*}{MiT-B3} & & 0.735 & 0.706 & 0.641 & 0.040 & 0.055 & 0.090 & 0.917 & 0.915 & 0.901 & 0.764 & 0.778 & 0.785 & 0.833 & 0.824 & 0.793 \\
                                & \checkmark & 0.748 & 0.716 & 0.642 & 0.035 & 0.052 & 0.083 & 0.923 & 0.919 & 0.906 & 0.785 & 0.794 & 0.800 & 0.846 & 0.833 & 0.800 \\ \midrule
        \multirow{2}{*}{MiT-B5} & & 0.763 & 0.737 & 0.675 & 0.037 & 0.057 & 0.085 & 0.931 & 0.930 & 0.920 & 0.814 & 0.825 & 0.829 & 0.860 & 0.850 & 0.822 \\
                                & \checkmark & 0.760 & 0.730 & 0.667 & 0.030 & 0.046 & 0.062 & 0.913 & 0.912 & 0.903 & 0.755 & 0.770 & 0.781 & 0.839 & 0.831 & 0.808 \\
        \bottomrule
    \end{tabular}
    \caption{Out-of-domain evaluation on the Foggy Cityscapes validation sets ($\beta = 0.005$, $\beta = 0.01$, $\beta = 0.2$) using various backbone sizes for DeepLabV3+ and SegFormer. Models are trained on Cityscapes and tested without re-training, allowing assessment of reliability and robustness under increasing fog density.}
    \label{tab:od_evaluation}
\end{table*}

\textbf{Out-of-Domain Evaluation.} The results in Table~\ref{tab:od_evaluation} show the robustness of all models under increasing fog intensities on Foggy Cityscapes. As expected, segmentation performance (mIoU) consistently declines with stronger fog due to the growing domain gap. While CutMix has limited impact on segmentation robustness, it helps to maintain better calibration (ECE) and uncertainty quality (p(\text{acc}$|$\text{cer}) and p(\text{unc}$|$\text{inacc})) across most configurations. Notably, SegFormer shows higher robustness than DeepLabV3+, particularly with larger backbones (MiT-B3 and MiT-B5), where both mIoU and calibration degrade less severely under adverse conditions. This trend suggests that transformer-based architectures generalize more gracefully across domain shifts compared to CNN-based ones. Overall, the RSS confirms that CutMix generally leads to more reliable calibration and uncertainty estimates across varying fog levels, even when segmentation accuracy itself remains mostly unchanged. These findings indicate that CutMix primarily enhances overall reliability and robustness rather than improving raw segmentation performance.

\begin{figure*}[t!]
\begin{center}
    \includegraphics[width=1.0\textwidth]{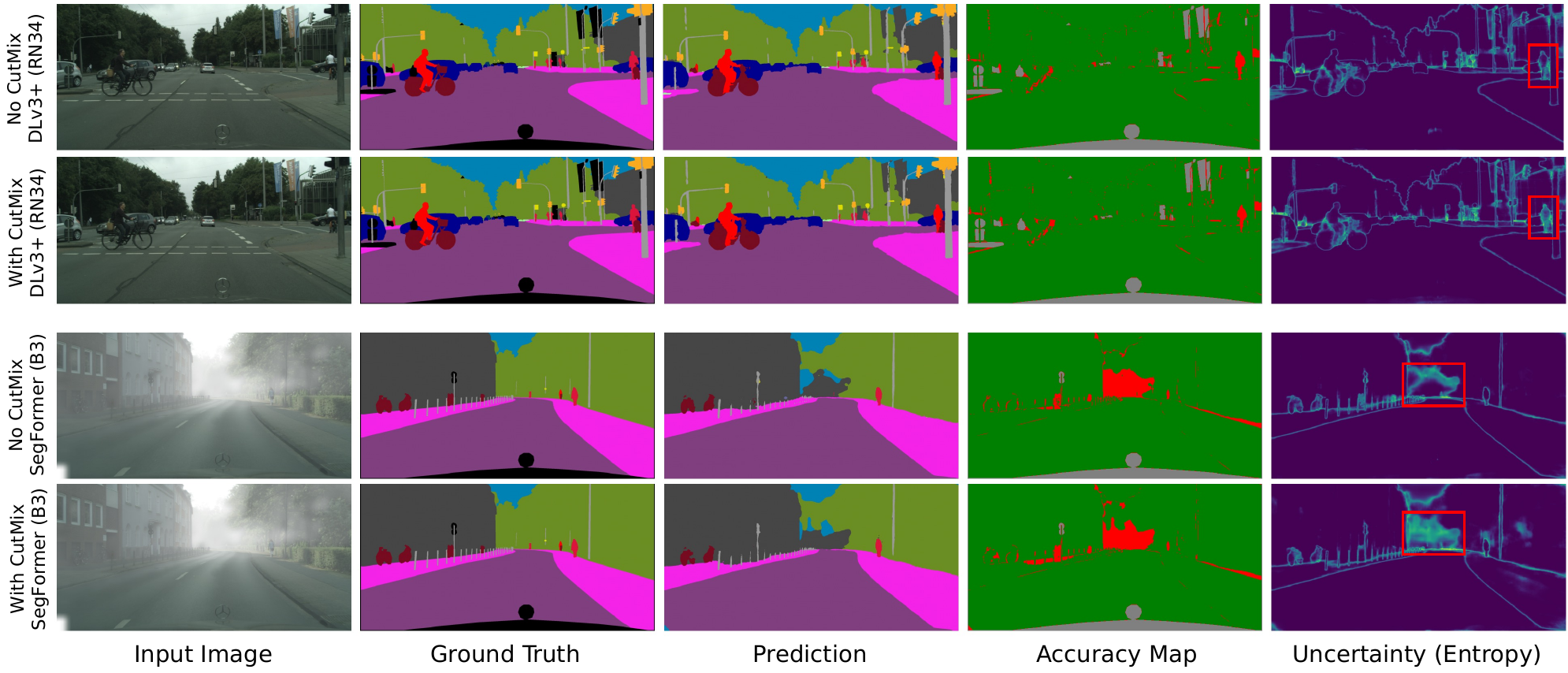}
    \caption{Qualitative examples of DeepLabV3+ (RN34) with and without CutMix on Cityscapes, and SegFormer (MiT-B3) on Foggy Cityscapes. The accuracy maps highlight correct predictions in green, incorrect ones in red, and classes ignored during training in gray.}
\label{fig: qualitative examples}
\end{center}
\end{figure*}

\textbf{Qualitative Evaluation.} 
Figure~\ref{fig: qualitative examples} compares qualitative results for DeepLabV3+ (RN34) with and without CutMix on Cityscapes and for SegFormer (MiT-B3) on Foggy Cityscapes. Across both architectures and datasets, segmentation predictions remain visually similar, confirming that CutMix does not substantially change the model’s class assignments. However, the CutMix-augmented variants exhibit higher uncertainty in regions corresponding to erroneous or ambiguous predictions, as highlighted by the red rectangles in the uncertainty maps. Overall, these qualitative observations corroborate the quantitative findings: while CutMix has limited effect on segmentation accuracy, it consistently improves the reliability, even under adverse conditions, ultimately making models more robustness as well.

\section{Conclusion}
This study systematically investigated the impact of CutMix on the accuracy, reliability, and robustness of semantic segmentation models. While CutMix is widely adopted, its effect on reliability and robustness in dense prediction tasks had not been considered yet. This is especially critical in light of recent findings by \citet{landgraf2025rethinking}, which revealed that semi-supervised semantic segmentation frameworks -- which use CutMix as a core component -- severely deteriorate the reliability. By isolating its effects from other components, we evaluated the influence of CutMix across in-domain and out-of-domain scenarios on two representative architectures: the CNN-based DeepLabV3+ and the transformer-based SegFormer. Our results reveal that CutMix exerts only a minor influence on segmentation accuracy and calibration but consistently improves uncertainty quality. These improvements persist under domain shifts, where CutMix-trained models demonstrate not only more reliable uncertainty estimates but also better calibration despite similar segmentation performance. In other words, our findings show that CutMix primarily enhances how models express their uncertainty rather than what they predict -- a crucial distinction for safety-critical applications.

This suggests that the reliability deterioration observed in semi-supervised segmentation frameworks \citep{landgraf2025rethinking} cannot be attributed to CutMix itself, but rather to other components such as pseudo-labeling or consistency regularization. Future work should investigate these interactions and test whether CutMix offers a general mechanism for enhancing reliability and robustness across tasks and modalities by extending evaluations to other domains, such as medical imaging or remote sensing. 

{
	\begin{spacing}{1.17}
		\normalsize
		\bibliography{ISPRSguidelines_authors} 
	\end{spacing}
}

\end{document}